\documentclass{article}

\usepackage{arxiv}

\usepackage[utf8]{inputenc} 
\usepackage[T1]{fontenc}    
\usepackage{hyperref}       
\usepackage{url}            
\usepackage{booktabs}       
\usepackage{amsfonts}       
\usepackage{nicefrac}       
\usepackage{microtype}      
\usepackage{lipsum}
\usepackage{graphicx}
\usepackage{threeparttable}
\usepackage{multirow}
\usepackage{booktabs}
\usepackage{amsmath}
\usepackage{verbatim}
\usepackage{graphicx}
\usepackage[caption=false,font=normalsize,labelfont=sf,textfont=sf]{subfig}
\usepackage{pifont}
\graphicspath{ {./images/} }

\title{K-Order Graph-oriented Transformer with GraAttention for 3D Pose and Shape Estimation}

\author{
 Weixi Zhao \\
  University of Chinese Academy of Sciences\\
  \texttt{zhaoweixi19@mails.ucas.edu.cn} \\
   \And
  Weiqiang Wang \\
  University of Chinese Academy of Sciences\\
  \texttt{wqwang@ucas.ac.cn} \\
}

\begin{document}
\maketitle
\begin{abstract}
We propose a novel attention-based 2D-to-3D pose estimation network for graph-structured data, named KOG-Transformer, and a 3D pose-to-shape estimation network for hand data, named GASE-Net. Previous 3D pose estimation methods focus on various modifications of the graph convolutional kernel, such as abandoning weight sharing or increasing the receptive field. Some of these methods employ attention-based non-local modules as auxiliary modules. In order to better model the relationship between nodes in graph-structured data and fuse the information of different neighbor nodes in a differentiated way, we make targeted modifications to the attention module and propose two modules designed for graph-structured data, graph relative positional encoding multi-head self-attention (GR-MSA) and K-order graph-oriented multi-head self-attention (KOG-MSA). By stacking GR-MSA and KOG-MSA, we propose a novel network KOG-Transformer\footnote{codes:\url{https://github.com/zhaoweixi/KOG-Transformer}} for 2D-to-3D pose estimation. Further, we propose a network for shape estimation on hand data, called GraAttention Shape Estimation Network (GASE-Net), which takes a 3D pose as input and gradually models the shape of a hand from sparse to dense.
We have empirically shown the superiority of KOG-Transformer through extensive experiments. The experimental results show that the KOG-Transformer significantly outperforms the previous state-of-the-art methods on the benchmark dataset Human3.6M. We evaluate the effect of GASE-Net on two hand datasets publicly available, ObMan and InterHand2.6M. The experimental results show that the GASE-Net can estimate the corresponding shapes for input poses with strong generalization ability.
\end{abstract}


\section{Introduction}
The 3D pose and shape estimation has attracted much attention in recent years from the computer vision community due to its numerous applications such as action recognition~\cite{weng2017spatio, yan2018spatial,li2019actional,jiang2021skeleton}, etc. 
The 2D-to-3D human pose estimation task aims to convert 2D coordinates of joints in images into corresponding 3D coordinates in the physical world. It is challenging since less information is contained in 2D coordinates. Previous works ~\cite{martinez2017simple, mehta2017monocular} have shown that the 2D coordinates plus connection structure information are vital to learning feature representations for 3D pose estimation. 
However, the CNN-based method~\cite{mehta2017monocular} lacks the ability to model the positional relationship between nodes for graph-structured data.

To better model the positional relationship between nodes, previous works ~\cite{doosti2020hope, zhao2019semantic, ci2019optimizing,liu2020comprehensive, xu2021graph} utilize the topological relationships modeled by graph convolutional networks (GCNs) to represent the positional relationships of nodes.
These GCN-based methods can be divided into two types: the methods focusing on improving graph convolutional kernels~\cite{zhao2019semantic, ci2019optimizing, liu2020comprehensive} and the methods focusing on designing pose estimation network structures~\cite{doosti2020hope, xu2021graph, li2021hierarchical}. 
Although these GCN-based methods have achieved good performance, they face the problem that only the information of neighbor nodes in topological relations can be aggregated in a single shot graph convolution computation. Therefore, more graph convolutional layers need to be stacked to aggregate global information, which may lead to over-smoothing problem.

To increase the receptive field, methods~\cite{zhao2019semantic, zhao2022graformer} use the attention mechanism to achieve global information fusion. Zhao \textit{et al.}~\cite{zhao2019semantic} use the Non-local module to increase the receptive field; Zhao \textit{et al.}~\cite{zhao2022graformer} proposes the GraAttention module to aggregate global information.
In the work~\cite{zhao2022graformer}, a learnable adjacency matrix is used to model topological relationships, which can be regarded as a special positional encoding in essence.

The attention mechanism has an obvious effect on increasing the receptive field, but how to perform positional encoding is still a problem worth thinking about.
In this paper, we propose to use graph-structure based relative positional encoding to model the body joint relations. The graph relative positional encoding we use can be viewed as a simplified version of GRPE~\cite{park2022grpe}. Unlike the previous works~\cite{shaw2018self, dai2019transformer}, where the difference value of the node indexes is used as the relative distance, we take the topological distance between two nodes in the graph structure as the relative distance in our work. %
Since there are no loops in the graph structure in our task, the case of multiple paths between two nodes is ignored.
When computing attention, we add trainable relative position information to queries and values respectively. With relative positional encoding based on the graph structure, we preserve the original topology but re-model the relationships between nodes. We refer to the proposed graph relative positional encoding attention module as \textbf{GR-MSA}.

A common problem of the graph convolutional kernel in the 3D pose estimation task is that the receptive field of a single shot is small, while the Chebyshev graph convolution~\cite{defferrard2016convolutional} can increase the receptive field by calculating the high-order graph Laplacian and learn parameter matrices used for the features of different orders separately. Inspired by this different treatment of neighbor nodes of different orders, we propose a novel attention module for graph-structured data, called K-order graph-oriented multi-head self-attention (\textbf{KOG-MSA}). We construct a set of masks corresponding to neighbor nodes of different orders according to the original adjacency matrix. For example, the zero-order neighbor corresponds to the node itself, and the corresponding mask is a diagonal matrix. We use the fully connected layer to map the attention results of different orders, and obtain the final result through weighted summation, and the weights are obtained through training. Through the differentiated processing of neighbor nodes of different orders, the KOG-MSA significantly improves the effect of node information fusion.
By stacking GR-MSA and KOG-MSA modules, we obtain a novel network for 2D-to-3D pose estimation called  K-order graph-oriented Transformer (\textbf{KOG-Transformer}).

For 3D shape estimation, the previous methods~\cite{ge20193d, boukhayma20193d, zhou2020monocular,chen2021joint,lin2021end, lin2021mesh} have achieved excellent performance. Ge \textit{et al.} adopt a graph convolutional network to predict shape with image features as input. Later, the methods~\cite{boukhayma20193d, zhou2020monocular,chen2021joint} predict 3D poses and parameters of MANO model~\cite{romero2017embodied} based on image features to model 3D shape. Recently, the transformer-based shape estimation methods~\cite{lin2021end, lin2021mesh} have been proposed and achieved excellent performance. Different from the above methods that take images as input, we propose a lightweight and simple network, called GraAttention Shape Estimation Network(\textbf{GASE-Net}), which takes a 3D pose as input for hand shape estimation. 
The GASE-Net is a sequential structure that models shapes from sparse to dense.
The first layer of GASE-Net is a Chebyshev graph convolution~\cite{defferrard2016convolutional} layer, followed by 5 layers of mesh attention block. Each mesh attention block is composed of a GraAttention~\cite{zhao2022graformer} layer and a node-level upsampling. The upsampling is implemented by a fully connected layer. Finally a linear layer is used to map the features into 3D space.
It should be noted that GASE-Net only takes 3D poses as input, so it is more suitable for hand data with less surface details. The GASE-Net is used in the subsequent stage of KOG-Transformer to predict the corresponding 3D shapes for the 3D poses. Since the GASE-Net takes 3D poses as input, it can be combined with other 3D pose estimation methods.

We demonstrate the effectiveness of our method by conducting comprehensive evaluation experiments and ablation studies on standard 3D benchmark dataset. The experimental results show that the proposed network, KOG-Transformer outperforms the state-of-the-art methods on Human3.6M~\cite{ionescu2013human3}. In particular, for the 2D ground truth inputs, we achieve 14 first-place results out of 15 categories on Human3.6M. 
The proposed KOG-Transformer is task-independent and thus can be easily applied to other graph regression tasks. 
The experimental results show that GASE-Net achieves high accuracy on two public hand datasets, ObMan and InterHand2.6M, and can predict the corresponding 3D shape for input poses with errors. GASE-Net can also be generalized to other objects with relatively flat surfaces, such as car shape estimation.

The contribution of this paper can be summarized in three aspects.
First, we propose to use graph relative positional encoding in the attention module, i.e. GR-MSA, to model the relation between the body joints. This relationship preserves the original topology information, while effectively affecting the information interaction between nodes.
Second, we propose a novel attention architecture, called KOG-MSA. This new attention architecture differentially fuses the information of neighbor nodes of different orders, which significantly improves the fusion effect.
By stacking the GR-MSA layer and KOG-MSA layer, we propose a novel network called KOG-Transformer, which can fuse the information of neighbor nodes of different orders in a differentiated manner while ensuring a large receptive field, achieves better performance on 2D-to-3D pose estimation task.
Third, we propose a lightweight shape estimation method called GASE-Net with a 3D pose as input, which effectively predicts 3D shape through a sparse-to-dense modeling approach with good generalization ability.

\section{Related Works}

\subsection{2D-to-3D Pose Estimation}
Early 2D-to-3D pose estimation methods~\cite{chen20173d, simon2017hand} mainly consider the geometric relationship of skeleton and joints.
Chen \textit{et al.}~\cite{chen20173d} propose to match 2D coordinates with a 3D pose database to yield 3D pose. 
Simon \textit{et al.}~\cite{simon2017hand} predict 3D pose based on the geometric relationship of the 2D coordinates of multiple views.
After that, data-driven methods are widely used.
Martinez \textit{et al.}~\cite{martinez2017simple} propose a linear layers based method, consisting of linear layers, batch normalization, dropout, and ReLU activation function, to predict 3D coordinates.
Hossain \textit{et al.}~\cite{hossain2018exploiting} take a sequence of consecutive 2D coordinates as input and utilize spatiotemporal information to predict 3D pose.
Later GCN-based methods~\cite{zhao2019semantic, doosti2020hope, ci2019optimizing, xu2021graph, zou2021modulated} become mainstream and achieve advanced results.
Zhao \textit{et al.}~\cite{zhao2019semantic} propose semantic graph convolution, which constructs a new affinity matrix using a learnable mask and the original adjacency matrix, then use non-local modules to enhance interaction among 2D joints.
Ci \textit{et al.}~\cite{ci2019optimizing} abandon the weight-sharing mode in GCN and instead build filters according to the skeleton structure.
Doosti \textit{et al.}~\cite{doosti2020hope} propose a U-shaped network that can perform trainable upsampling and downsampling on graph-structured data to predict 3D pose.
Xu \textit{et al.}~\cite{xu2021graph} propose the graph hourglass network and adopt the SE Block~\cite{hu2018squeeze} to fuse features extracted from different layers of the network.
Zou \textit{et al.}~\cite{zou2021modulated} add a learnable weight to the adjacency matrix as a new affinity matrix of GCN filter.
Recently, Zhao \textit{et al.}~\cite{zhao2022graformer} propose a novel attention module, which has a larger receptive field than ordinary GCNs and can better integrate global information.
In this paper, we further explore the use of attention mechanisms in the task of 2D-to-3D pose estimation.


\subsection{3D Shape Estimation}
Ge \textit{et al.}~\cite{ge20193d} utilize stacked hourglass network to extract latent features. Then the latent features are used as the input of the graph convolutional network to directly predict vertex coordinates. This approach is limited by the receptive field of graph convolution operations. The methods ~\cite{boukhayma20193d, zhou2020monocular,chen2021joint} use ResNet to extract image features, then generate the shape, pose and view parameters of MANO for 3D shape estimation. Lin \textit{et al.}~\cite{lin2021end} propose a transformer-based method, which uses HRNet~\cite{sun2019deep} to extract image features as the queries and performs positional encoding by adding a template human mesh to the image feature vector by concatenating the image feature with the 3D joint and vertex coordinates. 
On the basis of ~\cite{lin2021end}, Lin \textit{et al.} ~\cite{lin2021mesh} divide an input image into tokens and achieve better performance.
Different from the above methods, we propose a lightweight method that takes 3D poses as input to model relationships between vertices and predict shapes in a sparse-to-dense manner.

\subsection{Positional Encoding in Transformer}
Vaswani \textit{et al.}~\cite{vaswani2017attention} employ sine and cosine functions to encode two components for each position of a one-dimensional sequence. 
Shaw \textit{et al.}~\cite{shaw2018self} propose to add learnable position information to keys and values in self-attention, and this position information vector corresponds to the relative position of the node in the one-dimensional sequence. 
Dai \textit{et al.}~\cite{dai2019transformer} add additional learnable location parameters when the self-attention module calculates the dot product of queries and keys, and do not use the additional location information when calculating the values. 
Raffel \textit{et al.}~\cite{raffel2020exploring} simplify the setting of the position encoding by adding a learnable positional encoding vectors to queries and keys, but do not add to values. And each relative position distance corresponds to a positional encoding vector. 
He \textit{et al.} ~\cite{he2020deberta} add two additional items to the dot product result of queries and keys, which correspond to the relative positional relationship between the two nodes. 
Park \textit{et al.} ~\cite{park2022grpe} proposed to add the information corresponding to the spatial position and edge of the node to the result of the dot product of the queries and the keys, and this information corresponds to the relative position of the node, which is used to implement modeling on graph-structured data.
In this paper, we will explore how to design relative positional encoding to make the attention mechanism more suitable for graph-structured data.


\section{Methodology}
\begin{figure*}[htbp]
\centering
\includegraphics[scale=0.5]{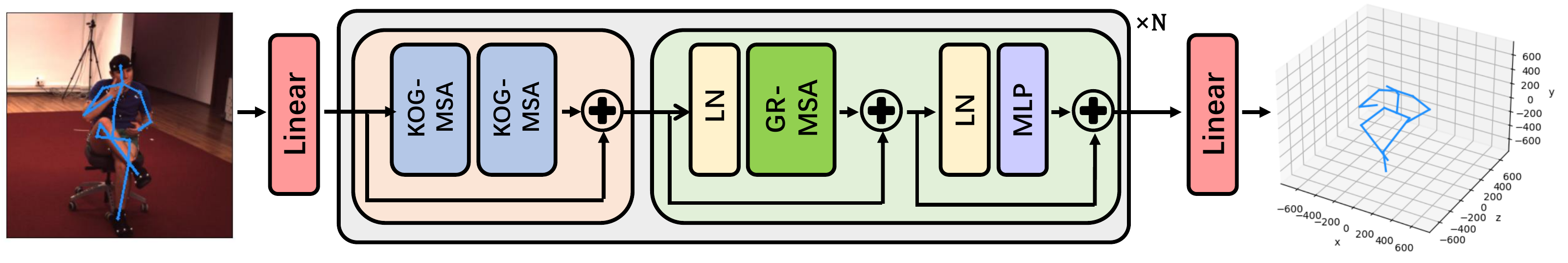}
\caption{Framework of KOG-Transformer. The core part is the stack of KOG-MSA block and GR-MSA layer, which boosts performance
for 2D-to-3D pose estimation tasks by exploiting relations among 2D joint}
\label{fig:pipeline}
\end{figure*}
As shown in Fig.~\ref{fig:pipeline}, the proposed KOG-Transformer takes 2D coordinates of joints as inputs and predicts 3D poses as a target. It is built by repeatedly stacking KOG-Transformer layers, each KOG-Transformer layer consists of a KOG-MSA block and a GR-MSA layer.
More details are given in subsections ~\ref{subsec: kog-msa}, ~\ref{subsec: gr-msa}. The GASE-Net is mainly built by stacking mesh attention blocks, and the details are given in subsection ~\ref{subsec: gase}.

\subsection{Preliminaries}
The gains achieved by the transformer-based methods ~\cite{lin2021end} mainly come from integrating the global information from a sequence. Concretely, basic attention with positional encoding\cite{vaswani2017attention} is calculated as:
\begin{equation}
   \begin{cases}
	\boldsymbol{q}_m=\left( \boldsymbol{f}_m+\boldsymbol{p}_m \right) W_Q\\
	\boldsymbol{k}_n=\left( \boldsymbol{f}_n+\boldsymbol{p}_n \right) W_K\\
	\boldsymbol{v}_n=\left( \boldsymbol{f}_n+\boldsymbol{p}_n \right) W_V\\
	A=softmax\left( \frac{QK^T}{\sqrt{d}} \right)\\
	\boldsymbol{y}_m=\sum_{n=1}^l{a_{m,n}\boldsymbol{v}_n}\\
\end{cases},
\end{equation}
where $m,n$=$ 1,2,...,l$, $\boldsymbol{f}_m,\boldsymbol{f}_n\in R^{d}$ represent the $m$-th and $n$-th rows of feature $F\in R^{l\times d}$ and $\boldsymbol{p}_m,\boldsymbol{p}_n\in R^{d}$ represent $m$-th and $n$-th positional encoding vectors. $W_Q,W_K,W_V\in R^{d\times d}$ are trainable weight matrices.
$A\in R^{l\times l}$ is the attention score and $A$=$\left( a_{m,n} \right) _{l\times l}$.
$Q,K\in R^{l\times d}$ denote queries and keys.
$\boldsymbol{q}_m,\boldsymbol{k}_n,\boldsymbol{v}_n,\boldsymbol{y}_m\in R^{d}$ denote the $m$-th or $n$-th query, key, value and attention result respectively.


\subsection{K-Order Graph-oriented Multi-head Self-Attention}
\label{subsec: kog-msa}
As illustrated in Fig.~\ref{fig:pipeline}, in the application of 3D pose estimation of human bodies, the pose of the human body in images is represented by sixteen 2D coordinates of joints.
The input linear layer maps the 2D coordinates to feature $F$ with dimensions $d$. The feature $F$ is then updated by multiple KOG-Transformer layers. Finally the output linear layer maps feature $F$ to 3D coordinates.

\begin{figure}[htbp]
\centering
\includegraphics[scale=0.5]{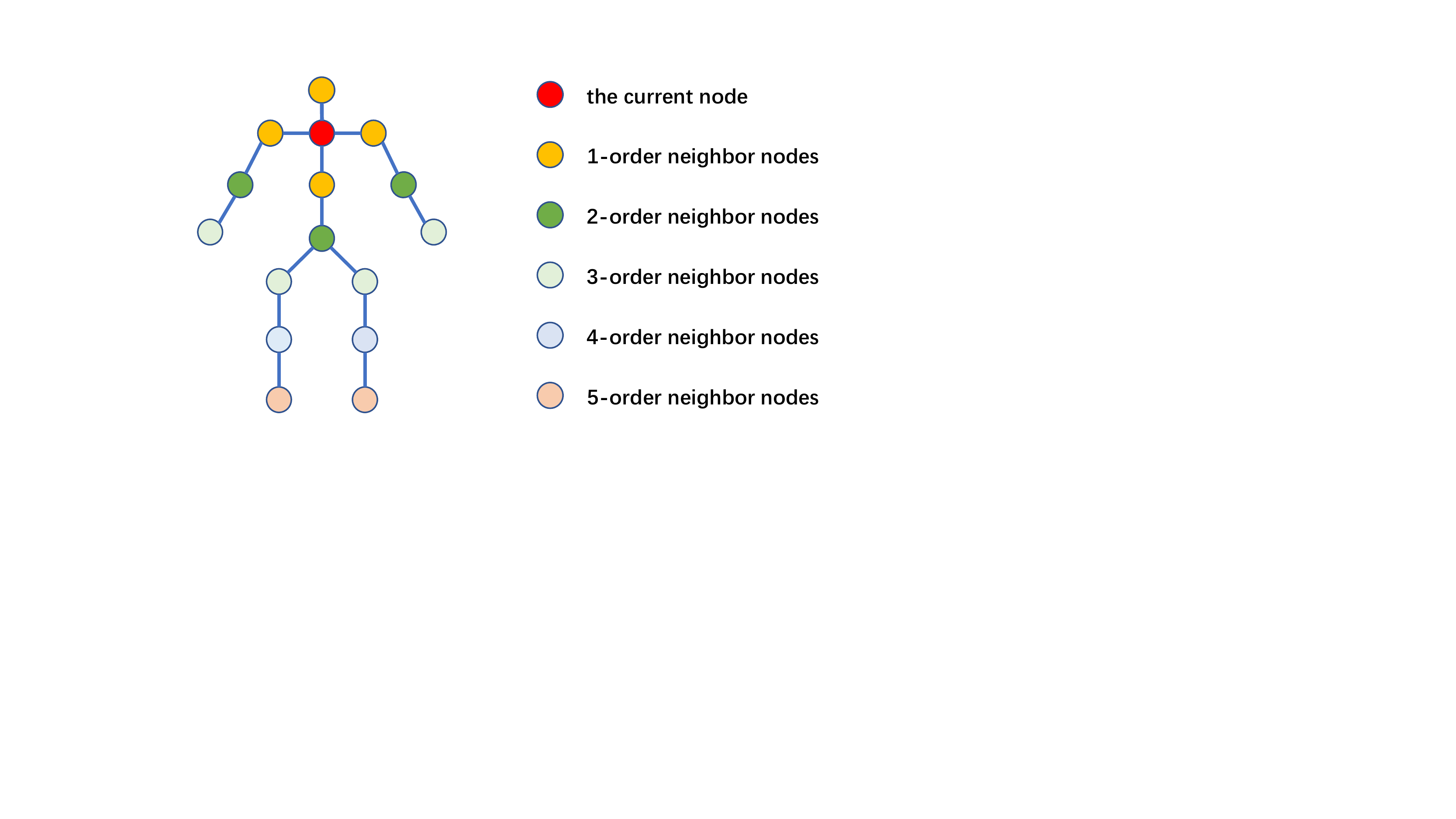}
\caption{Illustration of K-order neighbor nodes for a particular node.}
\label{fig: ill_ko}
\end{figure}

Inspired by Chebyshev graph convolution~\cite{defferrard2016convolutional,zhao2022graformer}, where aggregating node information by computing higher-order graph Laplacian matrices can achieve better results, we propose to compute the attention of the graph structure data independently for neighbor nodes of different order, and fuse the results adaptively. 
And we name it K-order graph-oriented multi-head self-attention(KOG-MSA).
Fig.~\ref{fig: ill_ko} shows an example of neighbor nodes of different orders.

\begin{figure}[htbp]
\centering
\includegraphics[scale=0.5]{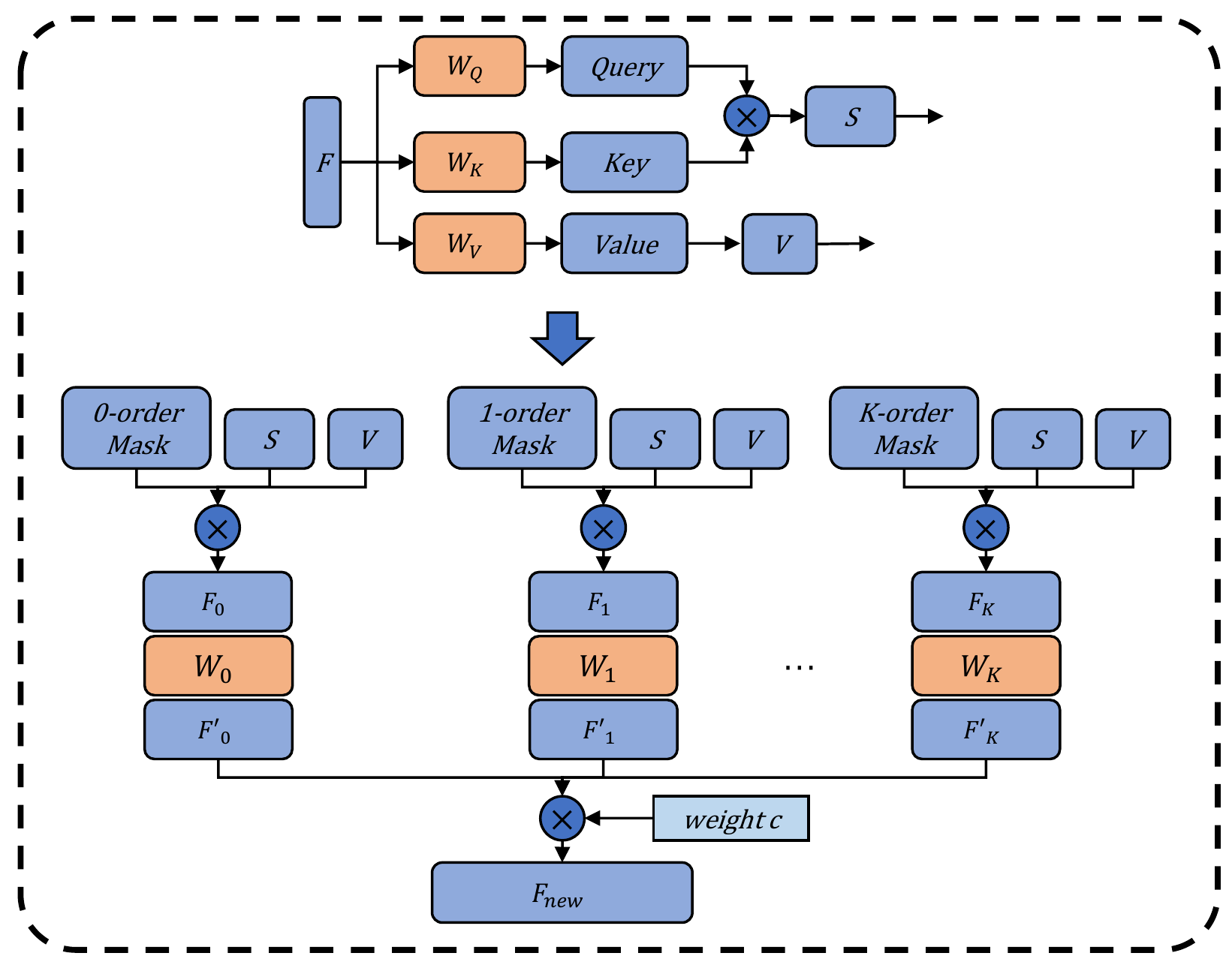}
\caption{Illustration of KOG-MSA.}
\label{fig: ko_sa}
\end{figure}

As shown in Fig.~\ref{fig: ko_sa}, for input graph features $F\in R^{l\times d}$, where $l$ is the number of nodes and $d$ is the feature dimension, we encode them as Query $Q$, Key $K$ and Value $V$ through three weight matrices $W_Q,W_K,W_V\in R^{d\times d}$, as:
\begin{equation}
    \begin{cases}
	    Q=F W_Q\\
	    K=F W_K\\
	    V=F W_V\\
	    S=Q{K}^{T} \\
    \end{cases},
\end{equation}
where $S\in R^{l\times l}$ is the attention value, and $S$ and $V$ are shared in subsequent computations of different orders. Then we calculate the information of neighbor nodes of different orders separately. The calculation for the $i$-th order neighbor node is as follows:

\begin{equation}
  \begin{cases}
	F_i=softmax\left( \frac{S+U^i}{\sqrt{d}} \right) V\\
	F_{i}^{'}=F_iW_i\\
\end{cases},
\end{equation}
where $F_i\in R^{l\times d}$, each row of $F_i$ corresponds to the fused feature of the $i$-th order neighbors of the node features, $W_i\in R^{d\times d}$ is used to re-represent the features of the $i$-th order neighbor nodes, $F'_i\in R^{l\times d}$ is the new feature of the $i$-th order neighbor nodes. The mask $U^i\in R^{l\times l}$ of order $i$ is defined as follows:
\begin{equation}
   u_{m,n}^{i}=\begin{cases}
	0&		,n\in \varOmega _{m}^{i}\\
	-\infty&		,else\\
\end{cases},
\end{equation}
where $\varOmega _{m}^{i}$ represents the set of the $i$-th order neighbor nodes of the $m$-th node. The value of order $i$ starts from zero to $K$, and we define the zero-order neighbor nodes of a node only consist of itself.
Finally, we use a trainable weight vector $c$=$[c_0,c_1,...,c_K]$ to perform a weighted summation of the features $F'_i$, as:
\begin{equation}
    F_{new}=\sum_{i=0}^K{c_i\times F'_i}.
\end{equation}
Note that $c$ in each KOG-MSA is different.
Through the above calculations, we utilize the topological structure of the graph in the attention module to achieve better node information fusion.

\subsection{Graph Relative Positional Encoding Multi-head Self-Attention}
\label{subsec: gr-msa}
Positional encoding in self-attention methods is used to represent the spatial positional information of tokens. However, the previous positional encoding methods~\cite{vaswani2017attention, shaw2018self, dai2019transformer, raffel2020exploring, he2020deberta} are mainly for sequence structured data, and is not suitable for graph data with topological structure.

\begin{figure}[htbp]
\centering
\includegraphics[scale=0.45]{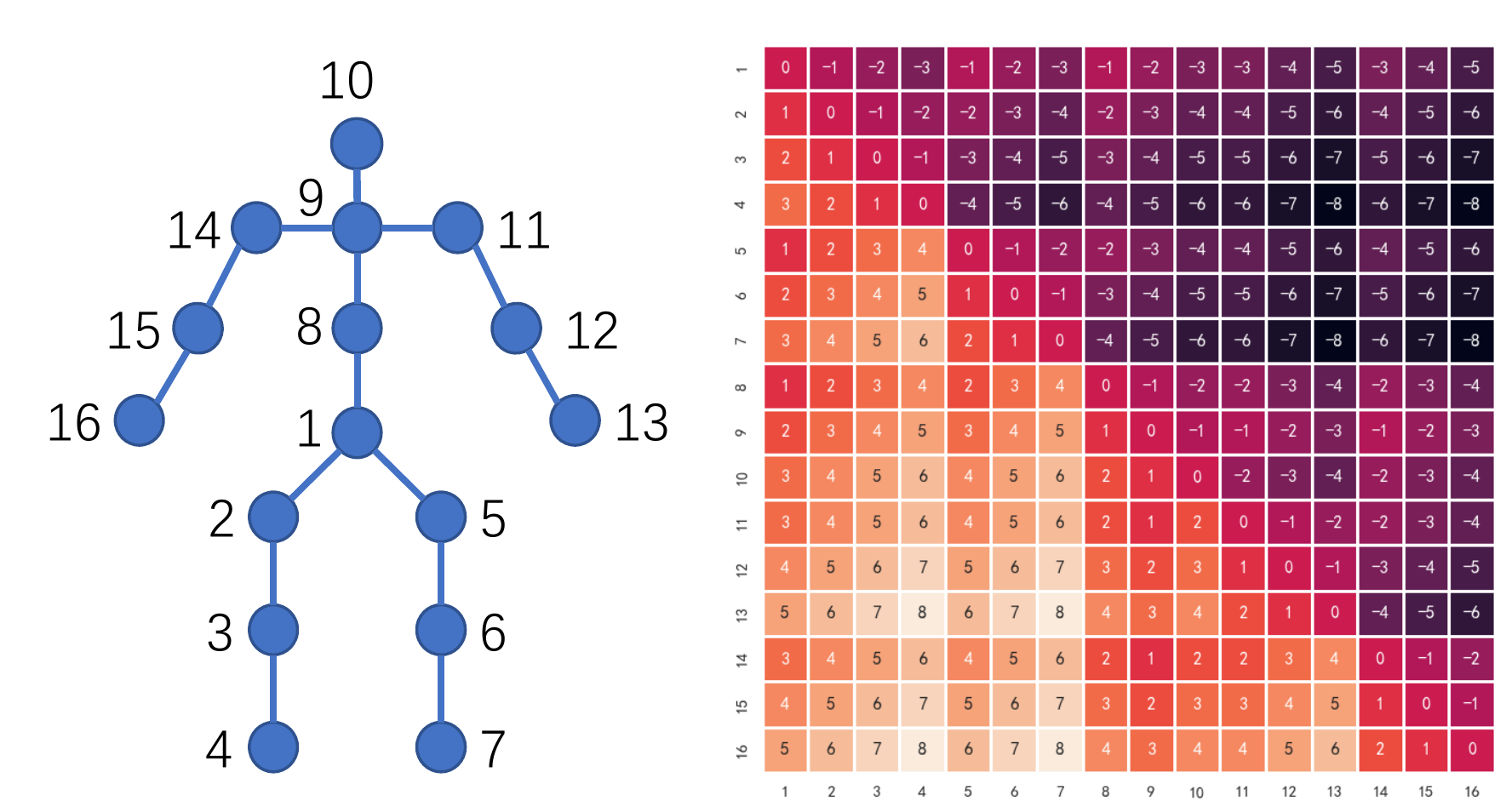}
\caption{Illustration of node indexes and distance matrix $H$ in GR-MSA.}
\label{fig: grpe}
\end{figure}

We propose a relative positional encoding for graph structures to enable self-attention for effective information fusion, called GR-MSA. The GR-MSA can be seen as a simplified version of GRPE~\cite{park2022grpe}, which only considers the node features of the graph data.
First, we still use three weight matrices $W_{Q}^{'},W_{K}^{'},W_{V}^{'}\in R^{d\times d}$ to encode the input features, as:

\begin{equation}
   \begin{cases}
	\boldsymbol{q}_m=\boldsymbol{f}_mW_{Q}^{'}\\
	\boldsymbol{k}_n=\boldsymbol{f}_nW_{K}^{'}\\
	\boldsymbol{v}_n=\boldsymbol{f}_nW_{V}^{'}\\
\end{cases},
\end{equation}
where $\boldsymbol{f}_m,\boldsymbol{f}_n,\boldsymbol{q}_m,\boldsymbol{k}_n,\boldsymbol{v}_n\in R^{d}$. Then we add learnable relative positional encoding to keys and values, as:
\begin{equation}
 \begin{cases}
	s_{m,n}=\boldsymbol{q}_m\left( \boldsymbol{k}_n+\boldsymbol{p}_{\gamma}^{K} \right) ^T\\
	A=softmax\left( S \right)\\
	\boldsymbol{f}_m=\sum_{n=1}^l{a_{m,n}\left( \boldsymbol{v}_n+\boldsymbol{p}_{\gamma}^{V} \right)}\\
\end{cases},
\end{equation}
where $S$=$\left(s_{m,n}\right)_{l\times l}$ and $S\in R^{l\times l}$ denotes the attention value matrix, $A$=$\left(a_{m,n}\right)_{l\times l}$ and $A\in R^{l\times l}$ denotes the attention score matrix,
$\boldsymbol{f}_m\in R^{d}$ is the new feature of node $m$. $P^K,P^V$ are trainable relative position vectors, $\boldsymbol{p}_{\gamma}^{K},\boldsymbol{p}_{\gamma}^{V}\in R^d$ are the $\gamma$-th row of $P^K,P^V$. $\gamma$ is obtained by the function $\varphi \left( m,n \right) $, which is the mapping function from indexes $m$ and $n$ to relative position vector index $\gamma$.
In the 2D-to-3D pose estimation task, we treat the skeleton graph as a undirected acyclic graph, and set the distance of each edge as 1. 
In GR-MSA, the $\varphi \left( m,n \right)$ function is defined as follows:
\begin{equation}
    \gamma =\varphi \left( m,n \right) =\min \left( \max \left( -\delta ,H_{m,n} \right) ,\delta \right) ,
\end{equation}
where $H\in R^{l\times l}$ is the relative distance matrix calculated from the skeleton structure, $H_{m,n}$ denotes the shortest path from node $m$ to node $n$. We define that the distance from a node with a small index to a node with a large index is negative; otherwise, the distance is positive. Fig.~\ref{fig: grpe} shows the corresponding index of graph nodes and the relative distance matrix $H$. For examples, $H_{8,7}$ is 4 and $H_{7,8}$ is -4. $\delta$ is an integer threshold for distance, and $P^K,P^V\in R^{\left( 2\delta +1 \right) \times d}$.
In implementation, to ensure that $\gamma$ is non-negative, we add an offset $\delta$ on it.
When the distance direction is ignored, i.e., all the distances are positive , the function $\varphi \left( m,n \right)$ is defined as:
\begin{equation}
    \varphi \left( m,n \right) =\min \left( H_{m,n}^{'},\delta \right) ,
\end{equation}
where $H'$=$\left( \left| H_{m,n} \right| \right) _{l\times l}$ is a symmetric matrix. In this case, $P^K,P^V\in R^{\left(\delta +1 \right) \times d}$. For GR-MSA, we use the directed distance for calculation. And we compare the effect of directed distance and undirected distance in the experiment.

\subsection{GraAttention-based Shape Estimation}
\label{subsec: gase}
The transition from 3D pose to 3D shape is a sparse-to-dense process. Different from 2D-to-3D pose estimation, which focuses on predicting depth information, the estimation from 3D pose to 3D shape expands the number of nodes in the same space, so it is more important to use the position relationship between nodes.

\begin{figure}[htbp]
\centering
\includegraphics[scale=0.45]{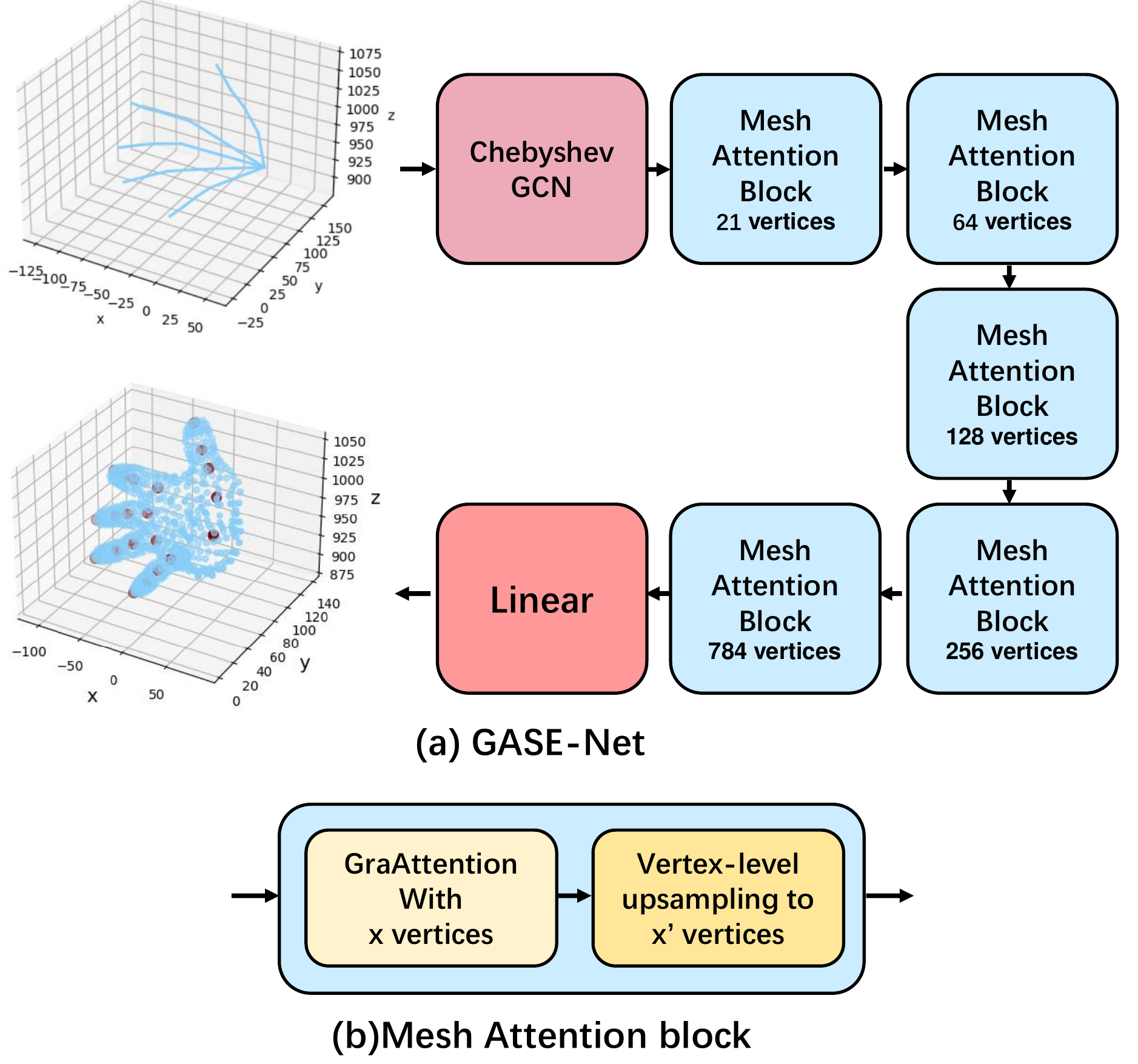}
\caption{Illustration of GASE-Net.}
\label{fig: gase}
\end{figure}

The entire GASE-Net is a sequential structure, as shown in Fig.~\ref{fig: gase}(a). The first layer is implemented by a graph convolutional layer, followed by five-layer MeshAttention blocks to model the mesh from sparse to dense. At last, a fully connected layer is used to map the features to a three-dimensional space.
We adopt Chebyshev graph convolution ~\cite{defferrard2016convolutional} as the first graph convolutional layer.
Each mesh attention block is implemented by a GraAttention~\cite{zhao2022graformer} layer and a vertex-level upsampling. The number of nodes in the learnable adjacency matrix of each GraAttention is set by a hyperparameter. 
In five mesh attention blocks, we set the number of nodes in a gradually increasing manner to achieve shape modeling from sparse to dense. After each GraAttention layer, a fully connected network is used to upsample at the node level to make the mesh denser, as shown in Fig.~\ref{fig: gase}(b). Finally, we employ a linear layer to map the mesh back into 3D space.

\subsection{Training}
To train the KOG-Transformer and GASE-Net, we apply a loss function to the final output to minimize the error between the 3D predictions and ground truth.
We use the mean squared errors (MSE) as the loss $L$ between 3D predictions and ground truth coordinates, i.e.,
\begin{equation}
   L=\frac{1}{\beta}\sum_{i=1}^{\beta}{\left( \left\| \tilde{J}_i-J_i \right\| ^2 \right)},
\end{equation}
where $\tilde{J}_{i}\in R^{l\times 3}$ denotes the predicted 3D coordinates, $J_{i}\in R^{l\times 3}$ is the 3D ground truth. For KOG-Transformer, $l$ is the number of joints. For GASE-Net, $l$ is the number of vertices. $\beta$ denotes the total number of training examples. $\left\|  \right\| $ denotes F-norm.


\section{Experiments}
In this section, we first introduce the experimental details and training settings. Next, we compare KOG-Transformer with other state-of-the-art methods and analyze the results. Then, we show the evaluation results of GASE-Net. Finally, we conduct ablation studies to evaluate the effectiveness of KOG-Transformer and GASE-Net.

\subsection{Dataset}
We adopt the most widely used benchmark dataset \textbf{Human3.6M}~\cite{ionescu2013human3} to evaluate the KOG-Transformer by comparing it with the state-of-the-art methods~\cite{zhao2022graformer,xu2021graph, ci2019optimizing, zhao2019semantic, hossain2018exploiting, fang2018learning, sun2017compositional}. We evaluate the model generalization ability by testing on the \textbf{MPI-INF-3DHP} \cite{mehta2017monocular} dataset.
We use two popular hand datasets, \textbf{ObMan}~\cite{hasson2019learning} and \textbf{InterHand2.6M}~\cite{moon2020interhand2} to evaluate the GASE-Net.

\textbf{Human3.6M}~\cite{ionescu2013human3} is a indoor environment dataset for the 3D human pose estimation task. It provides 3.6M accurate 3D poses captured by the MoCap system and  contains 15 actions performed by seven actors taken by four cameras.
There are two common evaluation protocols by splitting training and testing set in prior methods~\cite{martinez2017simple,zhao2019semantic, liu2020comprehensive,xu2021graph}.  
The first protocol uses subjects S1, S5, S6, S7 and S8 for training, and S9 and S11 for testing. Errors are calculated after the ground truth and predictions are aligned with the root joints.
The second protocol uses subjects S1, S5, S6, S7, S8 and S9 for training, and S11 for testing. 
We conduct experiments using the first protocol. In this way, there are 1,559,752 frames for training, and 543,344 frames for testing.

\textbf{MPI-INF-3DHP}~\cite{mehta2017monocular} is a 3D human body pose estimation dataset consisting of both constrained indoor and complex outdoor scenes. It records 8 actors performing 8 activities from 14 camera views. It consists over 1.3 million frames captured from the 14 cameras. We only use its test set to evaluate the generalization ability of the model.

\textbf{ObMan}~\cite{hasson2019learning}  is a synthetic dataset for hand-object interaction scenarios. In ObMan, the hand data is generated from MANO~\cite{romero2017embodied} and the object data is selected from the Shapenet~\cite{chang2015shapenet} dataset. The ObMan dataset contains 141,550 training frames and 6,285 evaluation frames. Each example contains an RGB image, a depth image, a 3D mesh of hand and object, and 3D coordinates for hand.

\textbf{Interhand2.6M}~\cite{moon2020interhand2} contains 687,548 single-hand frames and 673,514 interacting hand frames for training, 234,183 single-hand frames and 145,942 interacting hand frames for validation, and 488,968 single hand frames and 360,192 interacting hand frames for testing. The 3D shape of hands is denoted by MANO~\cite{romero2017embodied} parameters.

\subsection{Evaluation Metric}
For 3D pose estimation, we use the Mean-Per-Joint-Position-Error (MPJPE)~\cite{zhao2019semantic} in millimeters, which is calculated between the ground truth and the predicted 3D joints.
For 3D shape estimation, we use the Mean-Per-Vertex-Error (MPVE)~\cite{pavlakos2018learning} in millimeters, which measures the Euclidean distances between the ground truth vertices and the predicted vertices. Lower MPJPE and MPVE are better.

\subsection{Training Settings}
\paragraph{Pose Estiation}
For Human3.6M dataset, we normalize it according to~\cite{zhao2019semantic} before training and evaluation to alleviate the influence of multiple camera views. 
For KOG-Transformer, we set the number of N in Fig.~\ref{fig:pipeline} to 5 and adopt 4 heads for attention. The model dimension $d$ is set to 128 and with a dropout rate of 0.1 for GR-MSA, KOG-MSA and MLP.
We adopt Adam ~\cite{kingma2014adam} optimizer for optimization with an initial learning rate of 0.001 and mini-batches of 64. For Human3.6M, we multiply the learning rate by 0.9 every 50000 steps.

\paragraph{Shape Estimation}
The ObMan dataset provides 3D joints and vertex coordinates in the camera coordinate system. We simply convert the coordinates from meters to millimeters. The InterHand2.6M dataset provides the 3D joints and vertex coordinates(which are converted from MANO parameters) in the world coordinate system, we utilize the extrinsic matrix to transform the coordinates into the camera coordinate system. 
In experiments, the model dimension of GASE-Net are 32, and the dropout rate is 0.2.
For GASE-Net, we also adopt Adam ~\cite{kingma2014adam} optimizer for optimization with an initial learning rate of 0.00001 and mini-batches of 64. The learning rate decays by 0.8 per 5 epochs for InterHand2.6M and decays by 0.96 per 30 epochs for ObMan.

\begin{table*}[htb]
\caption{MPJPE (mm) results of KOG-Transformer on Human3.6M. This table is split into 2 groups. The inputs for the top group methods are images and the inputs for the bottom group are ground truth of 2D joints. (+) uses extra data from MPII~\cite{andriluka20142d}. }
	\centering
	\begin{threeparttable}
    	\resizebox{\textwidth}{30mm}{
         \begin{tabular}{l|ccccccccccccccc|c}
          \toprule 
	       Methods & Direct. & Discuss & Eating & Greet & Phone & Photo & Pose & Purch. & Sitting & SittingD. & Smoke & Wait & WalkD. & Walk & WalkT. & Avg. \\
	        \midrule 
            	Pavlakos~\cite{pavlakos2017coarse} CVPR'17 & 67.4 & 71.9 & 66.7 & 69.1 & 72.0 & 77.0 & 65.0 & 68.3 & 83.7 & 96.5 & 71.7 & 65.8 & 74.9 & 59.1 & 63.2 & 71.9\\
            	Metha~\cite{mehta2017monocular} 3DV'17 &52.6 & 64.1 & 55.2 & 62.2 & 71.6 & 79.5 & 52.8 & 68.6 & 91.8 & 118.4 & 65.7 & 63.5 & 49.4 & 76.4 & 53.5 & 68.6\\
            	Zhou~\cite{zhou2017towards} ICCV'17 & 54.8 & 60.7 & 58.2 & 71.4 & 62.0 & 65.5 & 53.8 & 55.6 & 75.2 & 111.6 & 64.1 & 66.0 & 51.4 & 63.2 & 55.3 & 64.9\\
            	Martinez~\cite{martinez2017simple} ICCV'17 & 51.8 & 56.2 & 58.1 & 59.0 & 69.5 & 78.4 & 55.2 & 58.1 & 74.0 & 94.6 & 62.3 & 59.1 & 65.1 & 49.5 & 52.4 & 62.9\\
            	Sun~\cite{sun2017compositional} ICCV'17(+)  & 52.8 & 54.8 & 54.2 & 54.3 & 61.8 & \textbf{53.1} & 53.6 & 71.7 & 86.7 & 61.5 & 67.2 & 53.4 & 47.1 & 61.6 & 53.4 & 59.1\\
            	Fang~\cite{fang2018learning} AAAI'18 & 50.1 & 54.3 & 57.0 & 57.1 & 66.6 & 73.3 & 53.4 & 55.7 & 72.8 & 88.6 & 60.3 & 57.7 & 62.7 & 47.5 & 50.6 & 60.4\\
            	Yang~\cite{yang20183d} CVPR'18(+) & 51.5 & 58.9 & 50.4 & 57.0 & 62.1 & 65.4 & 49.8 & 52.7 & 69.2 &     85.2 & 57.4 & 58.4 & \textbf{43.6} & 60.1 & 47.7 & 58.6\\
                Hossain~\cite{hossain2018exploiting} ECCV'18 & 48.4 & 50.7 & 57.2 & 55.2 & 63.1 & 72.6 & 53.0 & 51.7 & 66.1 & 80.9 & 59.0 & 57.3 & 62.4 & 46.6 & 49.6 & 58.3\\
                Zhao~\cite{zhao2019semantic} CVPR'19 & 48.2 & 60.8 & 51.8 & 64.0 & 64.6 & 53.6 & 51.1 & 67.4 & 88.7 & \textbf{57.7} & 73.2 & 65.6 & 48.9 & 64.8 & 51.9 & 60.8\\
                Ci ~\cite{ci2019optimizing} ICCV'19(+) & 46.8 & 52.3 & \textbf{44.7} & 50.4 & \textbf{52.9} & 68.9 & 49.6 & 46.4 & 60.2 & 78.9 & \textbf{51.2} & 50.0  & 54.8 & 40.4 & 43.3 & 52.7 \\
                Liu~\cite{liu2020comprehensive} ECCV'20 & 46.3 & 52.2 & 47.3 & 50.7 & 55.5 & 67.1 & 49.2 & 46.0 & 60.4 & 71.1 & 51.5 & 50.1 & 54.5 & 40.3 & 43.7 & 52.4 \\
                Xu ~\cite{xu2021graph} CVPR'21 & 45.2 & \textbf{49.9} & 47.5 & 50.9 & 54.9 & 66.1 & 48.5 & 46.3 & 59.7 & 71.5 & 51.4 & 48.6 & 53.9 & 39.9 & 44.1 & 51.9 \\
                Zhao~\cite{zhao2022graformer} CVPR'22 & 45.2 & 50.8 & 48.0 & \textbf{50.0} & 54.9 & 65.0 & 48.2 & 47.1 & 60.2 & 70.0 & 51.6 & 48.7 & 54.1 & 39.7 & 43.1 & 51.8\\
            \midrule 
            Ours & \textbf{45.0} & 50.0 & 45.8 & 50.4 & 53.8 & 62.7 & \textbf{47.8} & \textbf{45.6} & \textbf{58.7} & 66.2 & 51.4 & \textbf{48.0} & 53.0 & \textbf{38.4} & \textbf{41.3} & \textbf{50.5}\\
            \midrule 
                Martinez~\cite{martinez2017simple} (GT)& 37.7 & 44.4 & 40.3 & 42.1 & 48.2 & 54.9 & 44.4 & 42.1 & 54.6 & 58.0 & 45.1 & 46.4 & 47.6 & 36.4 & 40.4 & 45.5 \\
                
                Hossain~\cite{hossain2018exploiting} (GT) & 35.2 & 40.8 & 37.2 & 37.4 & 43.2 & 44.0 & 38.9 & 35.6 & 42.3 & 44.6 & 39.7 & 39.7 & 40.2 & 32.8 & 35.5 & 39.2 \\
                
                Zhao~\cite{zhao2019semantic} (GT) & 37.8 & 49.4 & 37.6 & 40.9 & 45.1 & 41.4 & 40.1 & 48.3 & 50.1 & 42.2 & 53.5 & 44.3 & 40.5 & 47.3 & 39.0 & 43.8 \\
                Zhou ~\cite{zhou2019hemlets} (GT) & 34.4 & 42.4 & 36.6 & 42.1 & 38.2 & 39.8 & 34.7 & 40.2 & 45.6 & 60.8 & 39.0 & 42.6 & 42.0 & 29.8 & 31.7 & 39.9 \\
                Ci ~\cite{ci2019optimizing} (GT)(+) & 36.3 & 38.8 & 29.7 & 37.8 & 34.6 & 42.5 & 39.8 & 32.5 & 36.2 & 39.5 & 34.4 & 38.4 & 38.2 & 31.3 & 34.2 & 36.3 \\
                Liu~\cite{liu2020comprehensive} (GT) & 36.8 & 40.3 & 33.0 & 36.3 & 37.5 & 45.0 & 39.7 & 34.9 & 40.3 & 47.7 & 37.4 & 38.5 & 38.6 & 29.6 & 32.0 & 37.8 \\
                Xu~\cite{xu2021graph} (GT) &  35.8 & 38.1 & 31.0 & 35.3 & 35.8 & 43.2 & 37.3 & 31.7 & 38.4 & 45.5 & 35.4 & 36.7 & 36.8 & 27.9 & 30.7 & 35.8\\
                Zhao~\cite{zhao2022graformer} (GT) & 32.0 & 38.0 & 30.4 & 34.4 & 34.7 & 43.3 & \textbf{35.2} & 31.4 & 38.0 & 46.2 & 34.2 & 35.7 & 36.1 & 27.4 & 30.6 & 35.2\\
            \midrule 
               Ours (GT) & \textbf{31.0} & \textbf{35.4} & \textbf{28.1} & \textbf{33.4} & \textbf{32.9} & \textbf{39.4} & 36.6 & \textbf{29.4} & \textbf{35.5} & \textbf{39.0} & \textbf{32.9} & \textbf{33.5} &  \textbf{34.5} & \textbf{26.8} & \textbf{28.6} & \textbf{33.2}\\
	        \bottomrule
    	\end{tabular}
    	}
	\end{threeparttable}
	
	\label{tab:human36m}
\end{table*}

\subsection{Performance and Comparison}
In this section, we evaluate the performance of KOG-Transformer on the benchmark dataset Human3.6M and compare it with other state-of-the-art methods. And we evaluate the performance of GASE-Net on two datasets, ObMan and InterHand2.6M.

\subsubsection{Performance of KOG-Transformer}


We conduct two groups of comparisons on the Human3.6M~\cite{ionescu2013human3} dataset in terms of differences in model inputs, in a manner widely adopted in previous works~\cite{zhao2022graformer, xu2021graph, zhao2019semantic}, as shown in TABLE.~\ref{tab:human36m}.
The input of the first group is the 2D keypoints detected from images. For our method, we adopt 2D coordinates detected by Cascaded Pyramid Network (CPN)~\cite{chen2018cascaded}. 
The results show that our method achieves the best performance on 7 actions.
In the second group, all methods take the same input, i.e. the ground-truth values of 2D coordinates. This group is a more fair comparison and reflects the upper bound of the model's performance. The results show that KOG-Transformer significantly outperforms other methods in this comparison, which shows the superiority of our method. It is important to note that all the methods involved in the comparison take only 2D coordinates as input and do not use additional information, such as camera parameters.

\subsubsection{Performence of GASE-Net}
We evaluate the performance of GASE-Net on two hand datasets ObMan and InterHand2.6M. We take the 3D poses and shape data of these two datasets as input and labels.
The results in TABLE~\ref{tab:hand_mesh_results} can be divided into two groups. The left group takes the ground-truth 3D pose with zero pose error as input and then computes the Mean-Per-Vertex-Error(MPVE) based on the predicted mesh vertices. The right group takes the erroneous 3D poses as input and computes Mean-Per-Vertex-Error based on the predicted mesh vertices. Both groups are evaluated using the same model, which only takes ground-truth 3D poses as input during training. The results show that the model trained on the ground-truth 3D pose is still able to perform effective estimation for 3D poses with errors, which reflects the generalization of GASE-Net.

\begin{table}[htb]
\caption{MPVE (mm) results of GASE-Net on ObMan and Interhand2.6M for 3D pose to shape estimation.}
\centering
	\begin{threeparttable}
	\scalebox{0.9}{
        \begin{tabular}{l|cc|cc}
          \toprule 
	       Dataset & MPVE & Pose MPJPE &  MPVE & Pose MPJPE\\
	        \midrule 
            	ObMan & 1.33 & 0 & 2.27 & 1.66 \\
            	InterHand2.6M & 1.56 & 0 & 13.05 & 13.11\\
	        \bottomrule
    	\end{tabular}
    	}
	\end{threeparttable}
\label{tab:hand_mesh_results}
\end{table}

\subsubsection{Generalization Ability}

To evaluate the generalization capabilities of KOG-Transformer, we train it on Human3.6M and evaluate it on the test set of MPI-INF-3DHP \cite{mehta2017monocular}.
The test set of MPI-INF-3DHP includes 3 settings, studio with green screen(GS), studio without green screen(noGS) and outdoors.
We use the same metric as ~\cite{mehta2017monocular}, including 3D Percentage of Correct Keypoints (3D PCK) with thresholds of 150 mm and the Area Under the Curve (AUC) computed for a range of PCK thresholds. 
TABLE.~\ref{tab:general} shows that the KOG-Transformer and GraphSH~\cite{xu2021graph} achieve the best PCK performance and KOG-Transformer achieves the second performance of the AUC index, which indicates that KOG-Transformer has good generalization performance. 

\begin{table}[htb]
\caption{Results on MPI-INF-3DHP test set.}
	\centering
	\begin{threeparttable}
	  \scalebox{0.7}{
        \begin{tabular}{l|c|ccccc}
          \toprule 
	        & Training data & GS & noGS & Outdoor & (PCK) & (AUC)\\
	        \midrule 
	            Martinez~\cite{martinez2017simple}  & H36M & 49.8 & 42.5 & 31.2 & 42.5 & 17.0 \\
	            Mehta~\cite{mehta2017monocular} &  H36M & 70.8 & 62.3 & 58.8 & 64.7 & 31.7\\
	            Yang~\cite{yang20183d} & H36M+MPII & - & - & - & 69.0 & 32.0\\
	            Zhou~\cite{zhou2017towards} & H36M+MPII & 71.1  & 64.7 & 72.7 & 69.2 & 32.5 \\
	            Luo~\cite{luo2018orinet} & H36M & 71.3 & 59.4 & 65.7 & 65.6 & 33.2\\
	            Ci~\cite{ci2019optimizing} & H36M & 74.8 & 70.8 & 77.3 & 74.0 & 36.7\\
	            Zhou~\cite{zhou2019hemlets} & H36M+MPII & 75.6 & 71.3 & 80.3 & 75.3 & 38.0\\
	            Xu~\cite{xu2021graph} & H36M & 81.5 & 81.7 & 75.2 & 80.1 & 45.8\\
	            Zhao~\cite{zhao2022graformer} & H36M & 80.1 & 77.9 & 74.1 & 79.0  & 43.8 \\
	       \midrule
	            ours & H36M & 81.8 & 78.2 & 77.5 & 80.1 & 44.0 \\
	        \bottomrule
    	\end{tabular}
        }
	\end{threeparttable}
	
	\label{tab:general}
\end{table}

For GASE-Net, due to the differences in the joint point order settings for 3D poses of different datasets and their settings for MANO model parameters, we do not evaluate the model generalization between different datasets. Instead, the generalization of the model is assessed on a single dataset through quantitative comparisons and qualitative visualization of the results.
TABLE~\ref{tab:hand_mesh_results} shows that when the model takes the erroneous 3D pose as input, the generated shape MPVE is close to the MPJPE of its input 3D pose. This shows that the relative position of the predicted 3D shape to the ground-truth 3D shape and the relative position of the input 3D pose to the ground-truth 3D pose is consistent.
The visualization results are shown in Fig~\ref{fig:mesh_vis_2}.

\subsection{Ablation Study}
\subsubsection{Discussion on model parameters}{ 
We start our ablation experiments by comparing the KOG-Transformer with other methods on the Human3.6M dataset, results are shown in TABLE~\ref{tab:params}. 
We report the results of our models of two configurations to show that our method can achieve better results with fewer parameters than other methods. %
KOG-Transformer has 5 layers, the feature dimension is 128 with a dropout rate of 0.1 and the order $K$ is 4. 
KOG-Transformer-mini has 5 layers, the feature dimension is 64 with a dropout rate of 0.1 and the order $K$ is 5. 
TABLE~\ref{tab:params} shows that KOG-Transformer significantly outperforms other methods. The lightweight version KOG-Transformer-mini not only has fewer parameters than the recent GraFormer~\cite{zhao2022graformer} and GraphSH~\cite{xu2021graph}, but also has significantly improved performance.
}

\begin{table}[htb]
	\centering
		\caption{Results on Human3.6M dataset under different parameter configurations.}
	\begin{threeparttable}
	\scalebox{0.8}{
        \begin{tabular}{l|cc|l|cc}
          \toprule 
	       Methods & Params & MPJPE & Methods & Params & MPJPE\\
	        \midrule 
	            GAT\cite{velivckovic2017graph} & 0.16M & 82.9 & 
	            FC\cite{martinez2017simple} & 4.29M & 45.5 \\
                ST-GCN\cite{yan2018spatial} & 0.27M & 57.4 &
                Pre-agg\cite{liu2020comprehensive} & 4.22M & 37.8 \\
                SemGCN\cite{zhao2019semantic} & 0.43M & 43.8 &
                GraphSH\cite{xu2021graph} & 3.70M & 35.8 \\
                GraFormer\cite{zhao2022graformer} & 0.65M & 35.2 &
                KOG-Transformer & 1.99M & 33.2
                \\
                KOG-Transformer-mini & 0.54M & 34.0 & & \\
                \bottomrule
    	\end{tabular}
        }
	\end{threeparttable}

	\label{tab:params}
\end{table}

\subsubsection{Effects of KOG-Transformer's Modules}{
We evaluate KOG-Transformer's modules on the benchmark dataset Human3.6M~\cite{ionescu2013human3}.
For GR-MSA, we analyze three factors through ablation experiments: the effect of different relative positional encoding methods, the effect of path direction and the effect of the maximum path threshold $\delta$.

First, we compare the performance of different positional encoding methods for the attention mechanism in the pose estimation task. In natural language processing tasks, many positional encoding methods for one-dimensional sequence data have been proposed. Such as, Sinusoidal positional encoding~\cite{vaswani2017attention}(Sinusoidal), learnable positional encoding~\cite{devlin2018bert}(Embedding), Relative position representations~\cite{shaw2018self}(Relative1) and ~\cite{dai2019transformer}(Relative2). For graph-structured data, GraAttention~\cite{zhao2022graformer} and graph relative positional encoding~\cite{park2022grpe} are proposed. In our method, we propose a new graph relative method, which can be seen as a simplified version of GRPE~\cite{park2022grpe} that is more suitable for 2D-to-3D pose estimation tasks. We evaluate the performance of GR-MSA by replacing it with the attention layer with the above positional encoding~\cite{vaswani2017attention, devlin2018bert,shaw2018self,dai2019transformer,zhao2022graformer}.
The results are shown in TABLE.\ref{tab:abl_grmsa}.
From the results, we can find that for the attention mechanism in pose estimation tasks, using relative positional encoding~\cite{shaw2018self, dai2019transformer} outperforms absolute positional encoding~\cite{vaswani2017attention,devlin2018bert}. By modifying the measure of relative position, our graph relative positional encoding outperforms the relative position measure~\cite{shaw2018self} based on node ordinal relationships.

\begin{table}[htb]
	\caption{GR-MSA ablation experiments of positional encoding.}
	\centering
	\begin{threeparttable}
		  \scalebox{0.9}{
        \begin{tabular}{lc|lc}
          \toprule 
	       Positional Encoding & MPJPE & Positional Encoding & MPJPE\\
	        \midrule
	        Sinusoidal~\cite{vaswani2017attention} & 57.1  & Embedding~\cite{devlin2018bert}  & 56.2\\
	         \midrule
	        Relative1~\cite{shaw2018self} & 34.8 & Relative2~\cite{dai2019transformer} & 43.5 \\
	        \midrule
            GraAttention~\cite{zhao2022graformer} & 37.5  & Ours  & 33.2 \\
	        \bottomrule
    	\end{tabular}
    	}
	\end{threeparttable}
	\label{tab:abl_grmsa}
\end{table}

Then, we experimentally evaluate the influence of the $\delta $ and influence of the direction of the relative position.
We compare the cases where the maximum threshold $\delta$ of the path is 1, 2, 3 and 4 for directed distance and 3, 5, 7 and 9 for undirected distance. The results are shown in TABLE~\ref{tab:abl_grmsa2}.
It should be noted that under this setting, the number of positional encoding vectors used in the directed and undirected path experiments in each column of the TABLE~\ref{tab:abl_grmsa2} is the same.
For example, the number of relative position vectors with a $\delta$ of 2 for a directed distance and a $\delta$ of 4 for an undirected distance is the same.

The results show that the use of directional representation is more helpful for the information aggregation of nodes. In the directed representation, the performance is best when the $\delta$ is 2, which means that when aggregating information, only the node itself and the first-order neighbor nodes have independent positional encoding vectors, and the second-order and above neighbor nodes share the positional encoding vector. This is also consistent with the previous method~\cite{ci2019optimizing} where local node relationships are considered very important. Furthermore, we find that the performance of directed and undirected distance representations is consistent with changes in the number of relative position vectors. This may indicate that there is an optimal value for the number of vectors, which encode position information, for a particular graph structure.

\begin{table}[htb]
	\caption{GR-MSA ablation experiments of $\delta$ on Human3.6M.}
	\centering
	\begin{threeparttable}
		  \scalebox{0.9}{
        \begin{tabular}{c|ccccc}
          \toprule 
	       threshold $\delta$ & 1 & 2 & 3 & 4\\
	      \midrule
	         MPJPE(directed) & 34.9 & 33.2 & 34.3 & 34.3 \\
	      \midrule
	        threshold $\delta$ & 2 & 4 & 6 & 8 \\
	      \midrule
	        MPJPE(undirected) & 43.4 & 42.8 & 43.7 & 44.0  \\
	        \bottomrule
    	\end{tabular}
    	}
	\end{threeparttable}
	\label{tab:abl_grmsa2}
\end{table}


For KOG-MSA, we evaluate its performance in two ways: we compare it with some GCN methods and compare its performance on different orders.
First, we replace the KOG-MSA layers in KOG-Transformer with different graph convolutional networks used in the previous 2D-to-3D pose estimation methods for comparison. Each graph convolutional network contains two graph convolutional layers, and each graph convolutional layer is followed by a 1D batch norm layer. And the GCN adopts residual connection. Specifically, we compare three graph convolutions, SemGCN~\cite{zhao2019semantic}, ModuGCN~\cite{zou2021modulated}, ChebGCN~\cite{defferrard2016convolutional,zhao2022graformer}. For ChebGCN, we set its order to 2 as ~\cite{zhao2022graformer}.
The results in TABLE~\ref{tab:abl_koga} show that our KOG-MSA, which is inspired by ChebGCN, achieves the best performance. Meanwhile, ChebGCN achieved second-place performance. We believe that this performance stems from increasing the receptive field of information fusion and differentially processing node information of different orders.

\begin{table}[htb]
	\caption{Comparison of KOG-MSA with different GCNs.}
	\centering
	\begin{threeparttable}
		  \scalebox{0.8}{
        \begin{tabular}{l|ccccc}
          \toprule 
	        Methods & KOG-MSA & ChebGCN~\cite{defferrard2016convolutional} & SemGCN~\cite{zhao2019semantic} & ModuGCN~\cite{zou2021modulated} \\
	        \midrule
	        MPJPE & 33.2 & 34.8 & 35.4 & 35.5 \\
	        \bottomrule
    	\end{tabular}
    	}
	\end{threeparttable}
	\label{tab:abl_koga}
\end{table}

Then we evaluate the performance of KOG-MSA for different orders $K$. For order $K$, KOG-MSA adopts neighbor nodes of order 0 to $K$ to aggregate new node information. The results in TABLE~\ref{tab:abl_koga2} show that 4-order achieves the best performance. 

\begin{table}[htb]
	\caption{Comparison of KOG-MSA with different order $K$.}
	\centering
	\begin{threeparttable}
		  \scalebox{0.9}{
        \begin{tabular}{l|ccccc}
          \toprule 
	       $K$ & 2 & 3 & 4 & 5 \\
	        \midrule
	        MPJPE & 36.2 & 34.1 & 33.2 & 35.2 \\
	        \bottomrule
    	\end{tabular}
    	}
	\end{threeparttable}
	\label{tab:abl_koga2}
\end{table}


For GASE-Net, we simply remove the GraAttention layers to compare the effects. After the removal, the remaining network contains a Chebyshev graph convolutional layer, learnable upsampling layers, and a linear layer. TABLE.~\ref{tab:ablation_shape} shows that GraAttention is effective for shape estimation.
\begin{table}[htb]
	\caption{Ablation experiments of GASE-Net on ObMan and Interhand2.6M with MPVE(mm) Metrics.}
	\centering
	\begin{threeparttable}
		  \scalebox{0.8}{
        \begin{tabular}{l|ccc}
          \toprule 
	        Methods &  ObMan & InterHand2.6M\\
	        \midrule 
	            GASE-Net   & 1.33 & 1.56 \\
	            ChebGCN$+$upsampling  & 3.42 & 4.53 \\
	        \bottomrule
    	\end{tabular}
    	}
	\end{threeparttable}

	\label{tab:ablation_shape}
\end{table}
}

\subsection{Visualization}{
Fig.~\ref{fig:kog_scale} shows the visualization of the trainable weights $c$ of KOG-MSA in a 5-layer 4-order KOG-Transformer.
For example, row 1-1 represents the weight $c$ in the first KOG-MSA module in the first layer of the KOG-Transformer, and columns 0 to 4 correspond to the values of $c_0$ to $c_4$ respectively.
It can be found from Fig.~\ref{fig:kog_scale} that for the first sublayer of each layer, the 1st and 2nd order neighbor nodes have a larger weight than other nodes. For the second sublayer of each layer, the zero-order nodes, the nodes themselves, have significantly less weight than other nodes. On the contrary, the weights of third-order neighbor nodes are significantly larger than other nodes. The above phenomenon shows that the first sub-layer of each layer pays more attention to the features of closer nodes, while the second sub-layer pays more attention to the features of farther nodes.

Fig.~\ref{fig:h36m_vis} shows the predicted 3D body results on Human3.6M~\cite{ionescu2013human3}. 
Fig.~\ref{fig:compare} shows the results of a comparison of our method with SemGCN\cite{zhao2019semantic} and GraFormer~\cite{zhao2022graformer} over two actions.

Fig.~\ref{fig:mesh_vis_2} shows the visualization results of GASE-Net using predicted 3D poses as inputs. For each dataset, we show three examples in three rows. In the first column, the orange skeleton is the 3D ground truth. In the first and second columns, the two blue skeletons are the same, and they are predicted by KOG-Transformer based on 2D ground truth. In the second and third columns, the green hands are the same which is predicted according to the blue skeleton. The orange hands in the third column are the 3D ground truth vertices. 
\begin{figure}[htbp]
\centering
\includegraphics[scale=0.40]{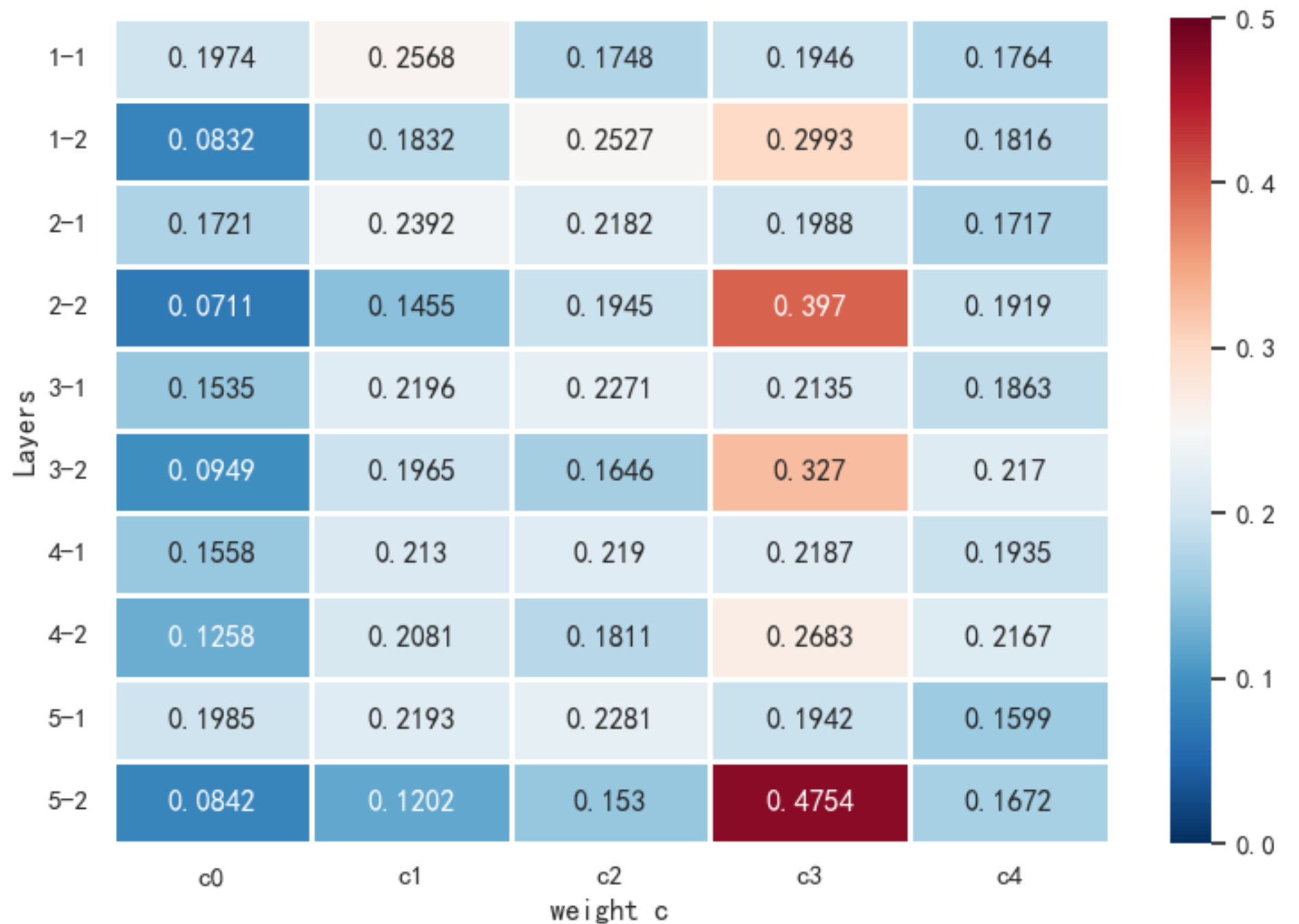}
\caption{
Visualization of weights $c$ in KOG-MSA modules in a 4-order KOG-Transformer. Each row represents a set of weights $c_0$ to $c_4$ in a KOG-MSA.
}
\label{fig:kog_scale}
\end{figure}

\begin{figure*}[htbp]
\centering
\includegraphics[scale=0.48]{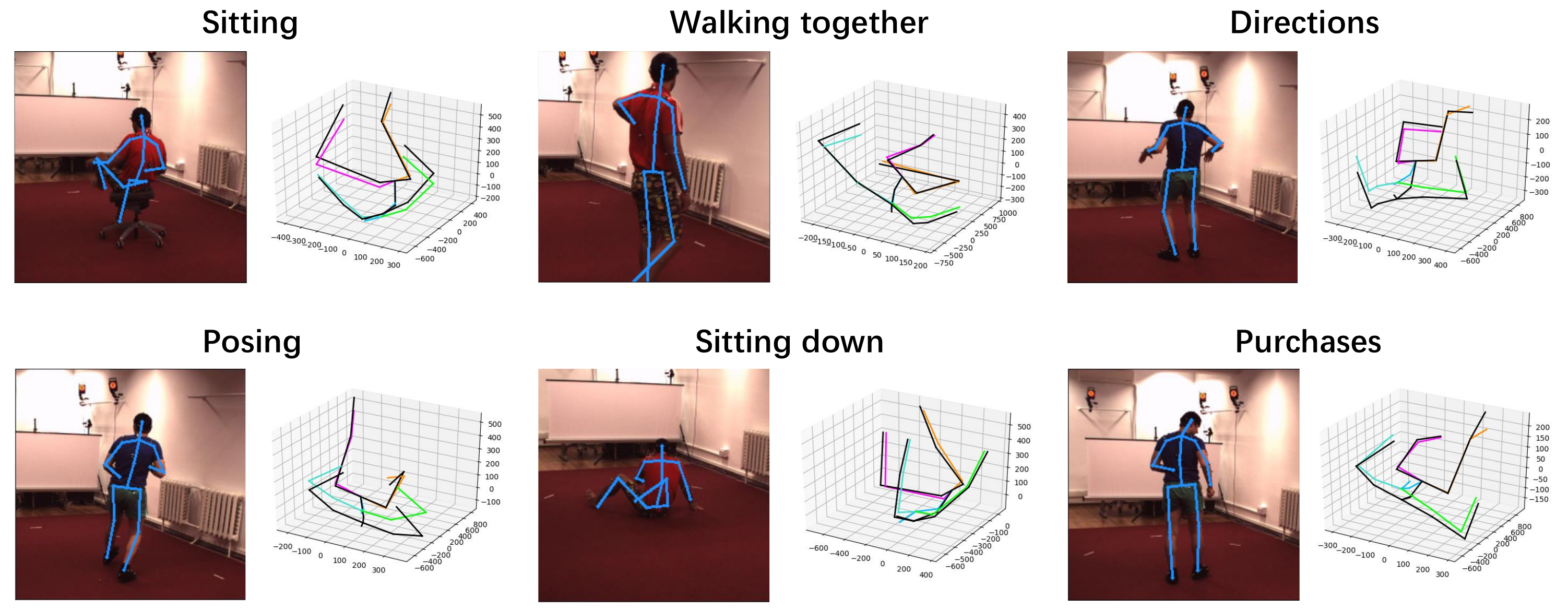}
\caption{The qualitative results of KOG-Transformer on Human3.6M~\cite{ionescu2013human3}. The 3D ground truth and 3D predictions are shown in black and color respectively.}
\label{fig:h36m_vis}
\end{figure*}

\begin{figure*}[htbp]
\centering
\includegraphics[scale=0.50]{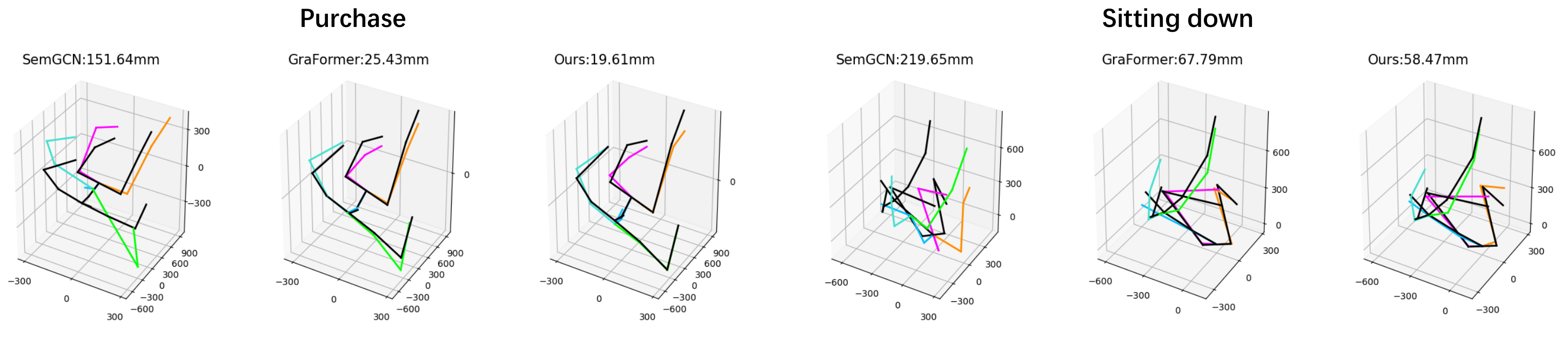}
\caption{Comparison with SemGCN~\cite{zhao2019semantic} and GraFormer~\cite{zhao2022graformer} on Human3.6M dataset. The 3D ground truth and 3D predictions are shown in black and color respectively.}
\label{fig:compare}
\end{figure*}

\begin{figure*}[htbp]
\centering
\includegraphics[scale=0.53]{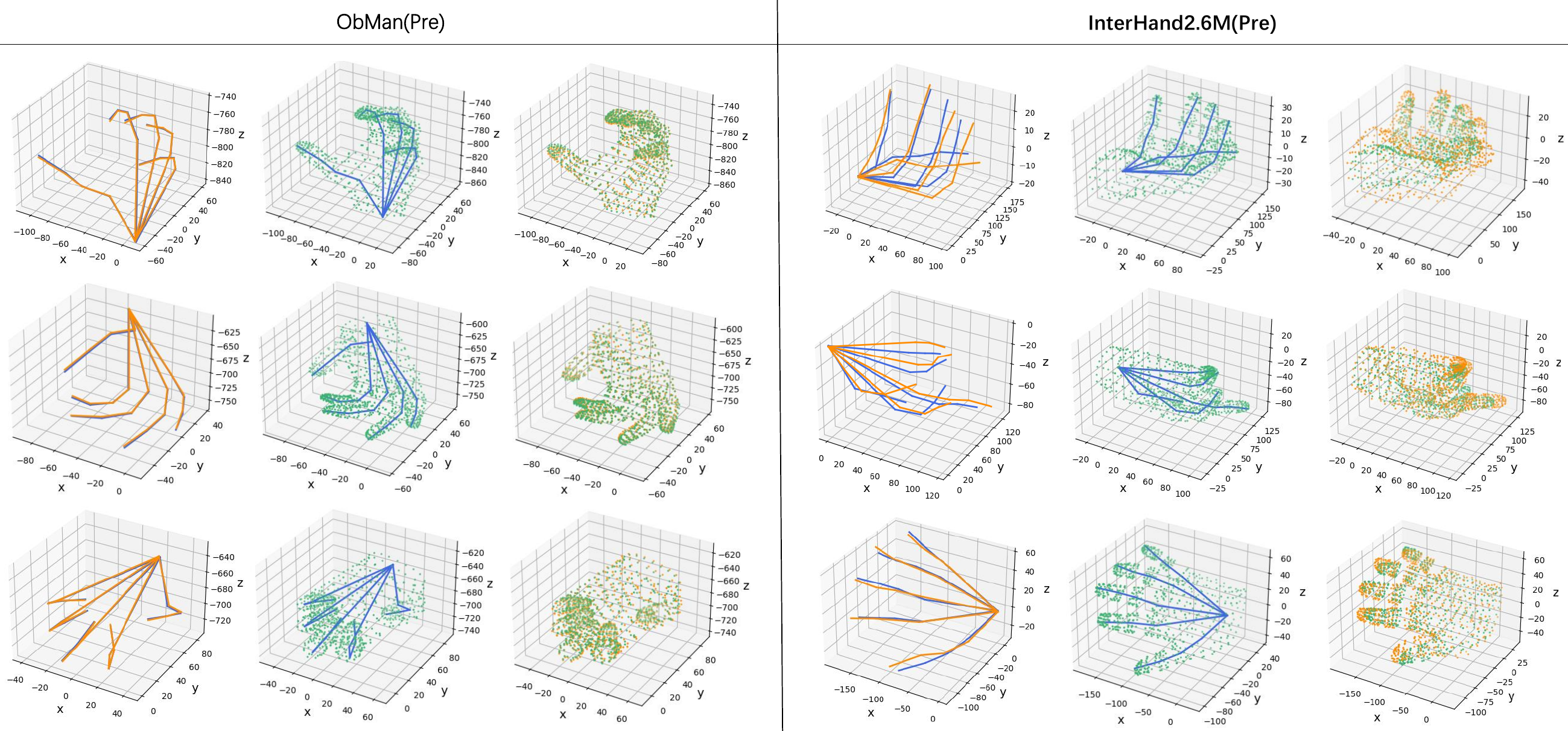}
\caption{Visualization of 3D pose and shape prediction on the ObMan and Interhand2.6M datasets.}
\label{fig:mesh_vis_2}
\end{figure*}
}

\section{Conclusions}
In this paper, we propose two attention modules, GR-MSA and KOG-MSA, for graph-structured data and a lightweight model GASE-Net for shape estimation. First, we experimentally demonstrate that a targeted modification of the positional encoding method can improve the performance of the attention module on pose estimation. Second, we verify that the differentiated aggregation of node information for graph-structured data can achieve better performance. Finally, our proposed simple and novel shape estimation method enables effective shape modeling in a sparse-to-dense manner.

\bibliographystyle{unsrt}  

\bibliography{references.bib}

\begin{thebibliography}{10}

\bibitem{weng2017spatio}
Junwu Weng, Chaoqun Weng, and Junsong Yuan.
\newblock Spatio-temporal naive-bayes nearest-neighbor (st-nbnn) for
  skeleton-based action recognition.
\newblock In {\em CVPR}, pages 4171--4180, 2017.

\bibitem{yan2018spatial}
Sijie Yan, Yuanjun Xiong, and Dahua Lin.
\newblock Spatial temporal graph convolutional networks for skeleton-based
  action recognition.
\newblock In {\em AAAI}, 2018.

\bibitem{li2019actional}
Maosen Li, Siheng Chen, Xu~Chen, Ya~Zhang, Yanfeng Wang, and Qi~Tian.
\newblock Actional-structural graph convolutional networks for skeleton-based
  action recognition.
\newblock In {\em CVPR}, pages 3595--3603, 2019.

\bibitem{jiang2021skeleton}
Songyao Jiang, Bin Sun, Lichen Wang, Yue Bai, Kunpeng Li, and Yun Fu.
\newblock Skeleton aware multi-modal sign language recognition.
\newblock In {\em CVPR}, pages 3413--3423, 2021.

\bibitem{martinez2017simple}
Julieta Martinez, Rayat Hossain, Javier Romero, and James~J Little.
\newblock A simple yet effective baseline for 3d human pose estimation.
\newblock In {\em ICCV}, pages 2640--2649, 2017.

\bibitem{mehta2017monocular}
Dushyant Mehta, Helge Rhodin, Dan Casas, Pascal Fua, Oleksandr Sotnychenko,
  Weipeng Xu, and Christian Theobalt.
\newblock Monocular 3d human pose estimation in the wild using improved cnn
  supervision.
\newblock In {\em 3DV}, pages 506--516. IEEE, 2017.

\bibitem{doosti2020hope}
Bardia Doosti, Shujon Naha, Majid Mirbagheri, and David~J Crandall.
\newblock Hope-net: A graph-based model for hand-object pose estimation.
\newblock In {\em CVPR}, pages 6608--6617, 2020.

\bibitem{zhao2019semantic}
Long Zhao, Xi~Peng, Yu~Tian, Mubbasir Kapadia, and Dimitris~N Metaxas.
\newblock Semantic graph convolutional networks for 3d human pose regression.
\newblock In {\em CVPR}, pages 3425--3435, 2019.

\bibitem{ci2019optimizing}
Hai Ci, Chunyu Wang, Xiaoxuan Ma, and Yizhou Wang.
\newblock Optimizing network structure for 3d human pose estimation.
\newblock In {\em ICCV}, pages 2262--2271, 2019.

\bibitem{liu2020comprehensive}
Kenkun Liu, Rongqi Ding, Zhiming Zou, Le~Wang, and Wei Tang.
\newblock A comprehensive study of weight sharing in graph networks for 3d
  human pose estimation.
\newblock In {\em ECCV}, pages 318--334. Springer, 2020.

\bibitem{xu2021graph}
Tianhan Xu and Wataru Takano.
\newblock Graph stacked hourglass networks for 3d human pose estimation.
\newblock In {\em CVPR}, pages 16105--16114, 2021.

\bibitem{li2021hierarchical}
Han Li, Bowen Shi, Wenrui Dai, Yabo Chen, Botao Wang, Yu~Sun, Min Guo, Chenlin
  Li, Junni Zou, and Hongkai Xiong.
\newblock Hierarchical graph networks for 3d human pose estimation.
\newblock 2021.

\bibitem{zhao2022graformer}
Weixi Zhao, Weiqiang Wang, and Yunjie Tian.
\newblock Graformer: Graph-oriented transformer for 3d pose estimation.
\newblock In {\em CVPR}, pages 20438--20447, 2022.

\bibitem{park2022grpe}
Wonpyo Park, Woong-Gi Chang, Donggeon Lee, Juntae Kim, et~al.
\newblock Grpe: Relative positional encoding for graph transformer.
\newblock In {\em ICLR2022 Machine Learning for Drug Discovery}, 2022.

\bibitem{shaw2018self}
Peter Shaw, Jakob Uszkoreit, and Ashish Vaswani.
\newblock Self-attention with relative position representations.
\newblock {\em arXiv preprint arXiv:1803.02155}, 2018.

\bibitem{dai2019transformer}
Zihang Dai, Zhilin Yang, Yiming Yang, Jaime Carbonell, Quoc~V Le, and Ruslan
  Salakhutdinov.
\newblock Transformer-xl: Attentive language models beyond a fixed-length
  context.
\newblock {\em arXiv preprint arXiv:1901.02860}, 2019.

\bibitem{defferrard2016convolutional}
Micha{\"e}l Defferrard, Xavier Bresson, and Pierre Vandergheynst.
\newblock Convolutional neural networks on graphs with fast localized spectral
  filtering.
\newblock {\em NeurIPS}, 29, 2016.

\bibitem{ge20193d}
Liuhao Ge, Zhou Ren, Yuncheng Li, Zehao Xue, Yingying Wang, Jianfei Cai, and
  Junsong Yuan.
\newblock 3d hand shape and pose estimation from a single rgb image.
\newblock In {\em CVPR}, pages 10833--10842, 2019.

\bibitem{boukhayma20193d}
Adnane Boukhayma, Rodrigo~de Bem, and Philip~HS Torr.
\newblock 3d hand shape and pose from images in the wild.
\newblock In {\em CVPR}, pages 10843--10852, 2019.

\bibitem{zhou2020monocular}
Yuxiao Zhou, Marc Habermann, Weipeng Xu, Ikhsanul Habibie, Christian Theobalt,
  and Feng Xu.
\newblock Monocular real-time hand shape and motion capture using multi-modal
  data.
\newblock In {\em CVPR}, pages 5346--5355, 2020.

\bibitem{chen2021joint}
Yujin Chen, Zhigang Tu, Di~Kang, Ruizhi Chen, Linchao Bao, Zhengyou Zhang, and
  Junsong Yuan.
\newblock Joint hand-object 3d reconstruction from a single image with
  cross-branch feature fusion.
\newblock {\em IEEE TIP}, 30:4008--4021, 2021.

\bibitem{lin2021end}
Kevin Lin, Lijuan Wang, and Zicheng Liu.
\newblock End-to-end human pose and mesh reconstruction with transformers.
\newblock In {\em CVPR}, pages 1954--1963, 2021.

\bibitem{lin2021mesh}
Kevin Lin, Lijuan Wang, and Zicheng Liu.
\newblock Mesh graphormer.
\newblock In {\em ICCV}, pages 12939--12948, 2021.

\bibitem{romero2017embodied}
Javier Romero, Dimitrios Tzionas, and Michael~J Black.
\newblock Embodied hands: Modeling and capturing hands and bodies together.
\newblock {\em ACM TOG}, 36(6):1--17, 2017.

\bibitem{ionescu2013human3}
Catalin Ionescu, Dragos Papava, Vlad Olaru, and Cristian Sminchisescu.
\newblock Human3. 6m: Large scale datasets and predictive methods for 3d human
  sensing in natural environments.
\newblock {\em IEEE TPAMI}, 36(7):1325--1339, 2013.

\bibitem{chen20173d}
Ching-Hang Chen and Deva Ramanan.
\newblock 3d human pose estimation= 2d pose estimation+ matching.
\newblock In {\em CVPR}, pages 7035--7043, 2017.

\bibitem{simon2017hand}
Tomas Simon, Hanbyul Joo, Iain Matthews, and Yaser Sheikh.
\newblock Hand keypoint detection in single images using multiview
  bootstrapping.
\newblock In {\em CVPR}, pages 1145--1153, 2017.

\bibitem{hossain2018exploiting}
Mir Rayat~Imtiaz Hossain and James~J Little.
\newblock Exploiting temporal information for 3d human pose estimation.
\newblock In {\em ECCV}, pages 68--84, 2018.

\bibitem{zou2021modulated}
Zhiming Zou and Wei Tang.
\newblock Modulated graph convolutional network for 3d human pose estimation.
\newblock In {\em ICCV}, pages 11477--11487, 2021.

\bibitem{hu2018squeeze}
Jie Hu, Li~Shen, and Gang Sun.
\newblock Squeeze-and-excitation networks.
\newblock In {\em CVPR}, pages 7132--7141, 2018.

\bibitem{sun2019deep}
Ke~Sun, Bin Xiao, Dong Liu, and Jingdong Wang.
\newblock Deep high-resolution representation learning for human pose
  estimation.
\newblock In {\em CVPR}, pages 5693--5703, 2019.

\bibitem{vaswani2017attention}
Ashish Vaswani, Noam Shazeer, Niki Parmar, Jakob Uszkoreit, Llion Jones,
  Aidan~N Gomez, {\L}ukasz Kaiser, and Illia Polosukhin.
\newblock Attention is all you need.
\newblock In {\em NeurIPS}, pages 5998--6008, 2017.

\bibitem{raffel2020exploring}
Colin Raffel, Noam Shazeer, Adam Roberts, Katherine Lee, Sharan Narang, Michael
  Matena, Yanqi Zhou, Wei Li, Peter~J Liu, et~al.
\newblock Exploring the limits of transfer learning with a unified text-to-text
  transformer.
\newblock {\em J. Mach. Learn. Res.}, 21(140):1--67, 2020.

\bibitem{he2020deberta}
Pengcheng He, Xiaodong Liu, Jianfeng Gao, and Weizhu Chen.
\newblock Deberta: Decoding-enhanced bert with disentangled attention.
\newblock {\em arXiv preprint arXiv:2006.03654}, 2020.

\bibitem{fang2018learning}
Hao-Shu Fang, Yuanlu Xu, Wenguan Wang, Xiaobai Liu, and Song-Chun Zhu.
\newblock Learning pose grammar to encode human body configuration for 3d pose
  estimation.
\newblock In {\em AAAI}, volume~32, 2018.

\bibitem{sun2017compositional}
Xiao Sun, Jiaxiang Shang, Shuang Liang, and Yichen Wei.
\newblock Compositional human pose regression.
\newblock In {\em ICCV}, pages 2602--2611, 2017.

\bibitem{hasson2019learning}
Yana Hasson, Gul Varol, Dimitrios Tzionas, Igor Kalevatykh, Michael~J Black,
  Ivan Laptev, and Cordelia Schmid.
\newblock Learning joint reconstruction of hands and manipulated objects.
\newblock In {\em CVPR}, pages 11807--11816, 2019.

\bibitem{moon2020interhand2}
Gyeongsik Moon, Shoou-I Yu, He~Wen, Takaaki Shiratori, and Kyoung~Mu Lee.
\newblock Interhand2. 6m: A dataset and baseline for 3d interacting hand pose
  estimation from a single rgb image.
\newblock In {\em ECCV}, pages 548--564. Springer, 2020.

\bibitem{chang2015shapenet}
Angel~X Chang, Thomas Funkhouser, Leonidas Guibas, Pat Hanrahan, Qixing Huang,
  Zimo Li, Silvio Savarese, Manolis Savva, Shuran Song, Hao Su, et~al.
\newblock Shapenet: An information-rich 3d model repository.
\newblock {\em arXiv preprint arXiv:1512.03012}, 2015.

\bibitem{pavlakos2018learning}
Georgios Pavlakos, Luyang Zhu, Xiaowei Zhou, and Kostas Daniilidis.
\newblock Learning to estimate 3d human pose and shape from a single color
  image.
\newblock In {\em CVPR}, pages 459--468, 2018.

\bibitem{kingma2014adam}
Diederik~P Kingma and Jimmy Ba.
\newblock Adam: A method for stochastic optimization.
\newblock {\em arXiv preprint arXiv:1412.6980}, 2014.

\bibitem{andriluka20142d}
Mykhaylo Andriluka, Leonid Pishchulin, Peter Gehler, and Bernt Schiele.
\newblock 2d human pose estimation: New benchmark and state of the art
  analysis.
\newblock In {\em CVPR}, pages 3686--3693, 2014.

\bibitem{pavlakos2017coarse}
Georgios Pavlakos, Xiaowei Zhou, Konstantinos~G Derpanis, and Kostas
  Daniilidis.
\newblock Coarse-to-fine volumetric prediction for single-image 3d human pose.
\newblock In {\em CVPR}, pages 7025--7034, 2017.

\bibitem{zhou2017towards}
Xingyi Zhou, Qixing Huang, Xiao Sun, Xiangyang Xue, and Yichen Wei.
\newblock Towards 3d human pose estimation in the wild: a weakly-supervised
  approach.
\newblock In {\em ICCV}, pages 398--407, 2017.

\bibitem{yang20183d}
Wei Yang, Wanli Ouyang, Xiaolong Wang, Jimmy Ren, Hongsheng Li, and Xiaogang
  Wang.
\newblock 3d human pose estimation in the wild by adversarial learning.
\newblock In {\em CVPR}, pages 5255--5264, 2018.

\bibitem{zhou2019hemlets}
Kun Zhou, Xiaoguang Han, Nianjuan Jiang, Kui Jia, and Jiangbo Lu.
\newblock Hemlets pose: Learning part-centric heatmap triplets for accurate 3d
  human pose estimation.
\newblock In {\em ICCV}, pages 2344--2353, 2019.

\bibitem{chen2018cascaded}
Yilun Chen, Zhicheng Wang, Yuxiang Peng, Zhiqiang Zhang, Gang Yu, and Jian Sun.
\newblock Cascaded pyramid network for multi-person pose estimation.
\newblock In {\em CVPR}, pages 7103--7112, 2018.

\bibitem{luo2018orinet}
Chenxu Luo, Xiao Chu, and Alan Yuille.
\newblock Orinet: A fully convolutional network for 3d human pose estimation.
\newblock {\em BMVC}, 2018.

\bibitem{velivckovic2017graph}
Petar Veli{\v{c}}kovi{\'c}, Guillem Cucurull, Arantxa Casanova, Adriana Romero,
  Pietro Lio, and Yoshua Bengio.
\newblock Graph attention networks.
\newblock {\em arXiv preprint arXiv:1710.10903}, 2017.

\bibitem{devlin2018bert}
Jacob Devlin, Ming-Wei Chang, Kenton Lee, and Kristina Toutanova.
\newblock Bert: Pre-training of deep bidirectional transformers for language
  understanding.
\newblock {\em arXiv preprint arXiv:1810.04805}, 2018.

\end{thebibliography}

\end{document}